\documentclass[10pt,twocolumn,letterpaper]{article}

\usepackage[american]{babel}
\usepackage{microtype}

\usepackage{cvpr}
\usepackage{times}
\usepackage{multirow}
\usepackage{graphicx}
\usepackage{amsmath}
\usepackage{amssymb}
\usepackage{tabularx}
\usepackage[noend]{algpseudocode}
\usepackage{algorithmicx,algorithm}
\usepackage[bold]{hhtensor}

\usepackage[pagebackref=true,breaklinks=true,colorlinks,bookmarks=false]{hyperref}

\cvprfinalcopy 

\def\cvprPaperID{3543} 

\ifcvprfinal\fi
\begin{document}

\title{Probabilistic End-to-end Noise Correction for Learning with Noisy Labels}
\author{Kun Yi \qquad\qquad Jianxin Wu\thanks{This research was partially supported by the National Natural Science Foundation of China (61772256, 61422203). J. Wu is the corresponding author.}\\
National Key Laboratory for Novel Software Technology\\
Nanjing University, Nanjing, China\\
{\tt\small yik@lamda.nju.edu.cn,\quad\tt\small wujx2001@nju.edu.cn}
}
\maketitle
\thispagestyle{empty}

\begin{abstract}
Deep learning has achieved excellent performance in various computer vision tasks, but requires a lot of training examples with clean labels. It is easy to collect a dataset with noisy labels, but such noise makes networks overfit seriously and accuracies drop dramatically. To address this problem, we propose an end-to-end framework called PENCIL, which can update both network parameters and label estimations as label distributions. PENCIL is independent of the backbone network structure and does not need an auxiliary clean dataset or prior information about noise, thus it is more general and robust than existing methods and is easy to apply. PENCIL outperforms previous state-of-the-art methods by large margins on both synthetic and real-world datasets with different noise types and noise rates. Experiments show that PENCIL is robust on clean datasets, too.
\end{abstract}

\section{Introduction}
Deep learning has shown very impressive performance on various vision problems, e.g., classification, detection and semantic segmentation. Although there are many factors for the success of deep learning, one of the most important is the availability of large-scale datasets with clean annotations like ImageNet~\cite{Deng2009CVPR_ImageNet}.

However, collecting a large scale dataset with clean labels is expensive and time-consuming. On one hand, expert knowledge is necessary for some datasets such as the fine-grained CUB-200~\cite{Wah2011CIT_CUB_200_2011}, which demands knowledge from ornithologists. On the other hand, we can easily collect a large scale dataset with noisy annotations through image search engines~\cite{Fergus2010PIEEE_CollectDataset,Krause2016ECCV_CollectDataset,Schroff2011TPAMI_CollectDataset}. These noisy annotations can be obtained by extracting labels from the surrounding texts or using the searching keywords~\cite{Xiao2015CVPR_Clothing1M}. For a huge dataset like JFT300M (which contains 300 million images), it is impossible to manually label it and inevitably about 20\% noisy labels exist in this dataset~\cite{Sun2017ICCV_JFT300M}. Hence, being able to deal with noisy labels is essential.

The label noise problem has been studied for a long time~\cite{Angluin1988ML_PreviousWork,Quinlan_1986ML_PreviousWork}. Along with the recent successes of various deep learning methods, noise handling in deep learning has gained momentum, too~\cite{Reed2015ICLR_Bootstrapping,Sukhbaatar2014_AN,Xiao2015CVPR_Clothing1M}. However, existing methods often have prerequisites that may not be practical in many applications, e.g., an auxiliary set with clean labels~\cite{Xiao2015CVPR_Clothing1M} or prior information about the noise~\cite{Patrini2017CVPR_Forward}. Some methods are very complex~\cite{Yao2018TIP_QualityEmbedding}, which hurts their deployment capability. Overfitting to noise is another serious difficulty. For a DNN with enough capacity, it can memorize the random labels~\cite{Zhang2017ICLR_NetworkMemory}. Thus, some noise handling methods may finally still overfit and their performance decline seriously, i.e., they are not robust. Their accuracies on the clean test set reach a peak in the middle of the training process, but will degrade afterwards and the accuracies after the final training epoch are poor~\cite{Patrini2017CVPR_Forward,Vahdat2017NIPS_CNNCRF}.

We attack the label noise problem from two aspects. First, we model the label for an image as a distribution among all possible labels~\cite{Gao2017TIP_DLDL} instead of a fixed categorical value. This \emph{probabilistic} modeling lends us the flexibility to handle noise-contaminated and noise-free labels in a unified manner. Second, inspired by~\cite{Tanaka2018CVPR_Daiki}, we maintain and update the label distributions in both network parameter learning (in which label distributions act as labels) and label learning (in which label distributions are updated to correct noise). Unlike~\cite{Tanaka2018CVPR_Daiki} which updates labels simply by using the running average of network predictions, we correct noise and update our label distributions in a principled \emph{end-to-end} manner. The proposed framework is called PENCIL, meaning \emph{probabilistic end-to-end noise correction in labels}. The PENCIL framework only uses the noisy labels to initialize our label distributions, then iteratively correct the noisy labels by updating the label distributions, and the network loss function is computed using the label distributions rather than the noisy labels.

Our contributions are as follows.%
\begin{itemize}
\item We propose an end-to-end framework PENCIL for noisy label handling. PENCIL is independent of the backbone network structure and does not need an auxiliary clean dataset or prior information about noise, thus it is easy to apply. PENCIL utilizes back-propagation to probabilistically update and correct image labels beyond updating the network parameters. To the best of our knowledge, PENCIL is the first method in this line.
\item We propose a variant of the DLDL method~\cite{Gao2017TIP_DLDL}, which is essential for correcting noise contained in our label distributions. PENCIL achieves state-of-the-art accuracy on datasets with both synthetic and real-world noisy labels (e.g., CIFAR-10, CIFAR-100 and Clothing1M).
\item PENCIL is robust. It is not only robust in learning with noisy labels, but also robust enough to apply in datasets with zero or small amount of \emph{potential} label noise (e.g., CUB-200) to improve accuracy.
\end{itemize}

\section{Related Works}

We first briefly introduce related works that inspired this work and other noise handling methods in the literature.

Deep label distribution learning was introduced in \cite{Gao2017TIP_DLDL} (called DLDL), which was proposed to handle label uncertainty by converting a categorical label (e.g., 25 years old) into a label distribution (e.g., a normal distribution whose mean is 25 and standard deviation is 3). The DLDL method uses constant label distributions and the Kullback-Leibler divergence to compute the network loss. In PENCIL, we use label distributions for a different purpose such that the label distributions can be updated and hence noise can be probabilistically corrected. The original DLDL method did not work in our setup and we designed a new loss function in PENCIL to overcome this difficulty.

For deep learning methods, \cite{Zhang2017ICLR_NetworkMemory} showed that a deep network with large enough capacity can memorize the training set labels even when they are randomly generated. Hence, they are particularly susceptible to noisy labels. Label noise can lead to serious overfitting and dramatically reduce network accuracy. However, \cite{Tanaka2018CVPR_Daiki} observed that when the learning rate is high, DNNs may maintain relatively high accuracy (i.e., the impact of label noise is not significant). This observation was utilized in~\cite{Tanaka2018CVPR_Daiki} to maintain an estimate of the labels using the running average of network predictions with a large learning rate. Then, these estimates were used as supervision signals to train the network. PENCIL is inspired by this observation and~\cite{Tanaka2018CVPR_Daiki}, too.

Label noise is an important issue and has long been researched~\cite{Angluin1988ML_PreviousWork,Quinlan_1986ML_PreviousWork}. There are mainly two types of label noise: symmetric noise and asymmetric noise, which are modeled in~\cite{Larsen1998ICAASSP_SN} and~\cite{Sukhbaatar2014_AN}, respectively. \cite{SurveyAboutNoise} is a survey of relatively early methods. \cite{Rolnick2017_DLRobustness} argued that deep neural networks are inherently robust to label noise to some extent. And, deep methods have achieved state-of-the-art results in recent years. Hence, we mainly focus on noise handling in deep learning models in this section.

One intuitive and easy solution is to delete all the samples which are considered as unreliable~\cite{Brodley1999JAIR_RemoveSample}. However, many difficult samples will be deleted, but these samples are important to algorithm's accuracy~\cite{Guyon1996KDD_RemoveSampleIsNotGood}. Thus, more profound noisy label handling methods become necessary. 

There are mainly two lines of attack to the the noisy label problem: constructing a special model based on noisy labels or using a robust loss function. The objective of these methods is to construct a noise-aware model which explicitly deals with noisy labels. \cite{Xiao2015CVPR_Clothing1M} constructed a model to deal with noisy labels, and tested their method on a real-world dataset collected by them. \cite{Vahdat2017NIPS_CNNCRF} proposed a framework called CNN-CRF, which combined convolutional neural networks (CNN) with conditional random fields (CRF) to characterize noisy labels. \cite{Yao2018TIP_QualityEmbedding} utilized similar ideas to determine the confidence of each label. This approach is gaining popularity in recent years (e.g., in~\cite{Lee2018CVPR_CleanNet,Ma2018ICML_LID,Wang2018CVPR_LOF}), and different techniques such as local inherent dimensionality have been brought into the noisy label learning domain.

Another effective approach is to design robust loss functions in order for a noise-tolerant model. Forward and backward methods~\cite{Patrini2017CVPR_Forward} explicitly modeled the noise transition matrix in loss computation. \cite{Ghosh2017AAAI_MSE} investigated the robustness of different loss functions, such as the mean squared loss, mean absolute loss and cross entropy loss. \cite{Zhang2018NIPS_Lq} combined advantages of the mean absolute loss and cross entropy loss to obtain a better loss function.

\cite{Tanaka2018CVPR_Daiki} did not fall in these two categories. It is special in the sense that it replaced the noisy label with their own estimate of the label (i.e., running average of the network's predictions). This approach is effective in noise handling but ad-hoc. PENCIL is partly inspired by this work, but more principled and effective.

Existing methods usually have prerequisites that are impractical, such as demanding an additional clean dataset (e.g., to curb overfitting) or a groundtruth noise transition matrix. When these prerequisites are not satisfied, they often fail to produce robust models. These methods are sometimes too complex to be deployed in real-world applications. In contrast, the proposed PENCIL method does not require additional information, and it can be easily applied to any backbone network.

\section{The Proposed PENCIL Method}

First of all, we define the notations for our study. Column vectors are denoted in bold (e.g., $\vec{x}$) and matrices in capital form (e.g., $X$). Specifically, $\vec{1}$ is a vector of all-ones. We use both hard labels and soft labels. The hard-label space is $\mathcal{H} = \{\vec{y}: \vec{y} \in \{0,1\}^c, \vec{1}^\top \vec{y} = 1 \}$, and the soft-label space is $\mathcal{S} = \{\vec{y}: \vec{y} \in [0,1]^c, \vec{1}^\top \vec{y} = 1 \}$. That is, a soft-label is a label distribution.

\subsection{Probabilistic modeling of noisy labels}

In a $c$-class classification problem, we have a training set $X = \{\vec{x}_1,\vec{x}_2,\dots,\vec{x}_n\}$. In the ideal scenario, every image $\vec{x}_i$ has a clean label $\vec{y}_i \in \mathcal{H}$, which is a one-hot vector (i.e., equivalent to an integer between 1 and $c$). In our noisy label problem, the labels might be wrong with relatively high probability and we use $\hat{\vec{y}}_i \in \mathcal{H}$ to denote labels which may contain noise. Using cross entropy, the loss function is
\begin{equation}
\mathcal{L} = -\frac{1}{n}\sum_{i=1}^n\sum_{j=1}^c \hat{y}_{ij}\log f_j(\vec{x}_i; \vec{\theta})\,,  \label{1} 
\end{equation}
where $\hat{y}_{ij}$ is the $j$'th element of $\hat{\vec{y}}_i$, $f$ is a model's prediction (processed by the softmax function) and $\vec{\theta}$ is the set of network parameters.

In PENCIL, we maintain a label distribution $\vec{y}_i^d \in \mathcal{S} = \{\vec{y}: \vec{y} \in [0,1]^c, \vec{1}^\top \vec{y} = 1 \}$ for every image $\vec{x}_i$, which is our estimate of the \emph{underlying noise-free} label for $\vec{x}_i$. $\vec{y}_i^d$ is used as the pseudo-groundtruth label in our learning, which is initialized based on the noisy label $\hat{\vec{y}}_i$. It is continuously updated (i.e., the noise is gradually corrected) through back-propagation. This probabilistic setting allows ample flexibility for noise correction. Note that our probabilistic modeling of the noisy labels is different from that in DLDL~\cite{Gao2017TIP_DLDL}. Label distributions in DLDL are fixed and cannot be updated.

In~\cite{Gao2017TIP_DLDL}, the loss function is KL-divergence:
\begin{gather}
\mathcal{L}  = \frac{1}{n}\sum_{i = 1}^n KL(\vec{y}_i^d || f(\vec{x}_i; \vec{\theta})), \text{and}  \label{2} \\
\mathrm{KL}(\vec{y}_i^d || f(\vec{x}_i; \vec{\theta})) = \sum_{j  = 1}^c y_{ij}^d \log\left(\frac{y_{ij}^d}{f_j(\vec{x}_i; \vec{\theta})}\right) \,.  \label{3}
\end{gather}
This loss is used in~\cite{Tanaka2018CVPR_Daiki}, too. However, KL-divergence is an asymmetric function. Hence, if we exchange the two operands in Eq.~\ref{2}, we obtain a new loss function
\begin{gather}
\mathcal{L}  = \frac{1}{n}\sum_{i = 1}^n KL(f(\vec{x}_i; \vec{\theta}) || \vec{y}_i^d), \text{and} \label{4}\\
\mathrm{KL}(f(\vec{x}_i; \vec{\theta}) || \vec{y}_i^d) = \sum_{j  = 1}^c f_j(\vec{x}_i; \vec{\theta}) \log\left(\frac{f_j(\vec{x}_i; \vec{\theta})}{y_{ij}^d}\right) \,.\label{5}
\end{gather}
We will soon show that Eq.~\ref{4} is more suitable for noise handling. In fact, Eq.~\ref{2} led to very poor results in our experiments and we propose to use Eq.~\ref{4} as one of the loss functions in PENCIL. More details will be discussed in Section~\ref{sec:class_loss}.

\subsection{End-to-end noise correction in labels}

\begin{figure}[t]
\begin{center}
\includegraphics[width=0.95\linewidth]{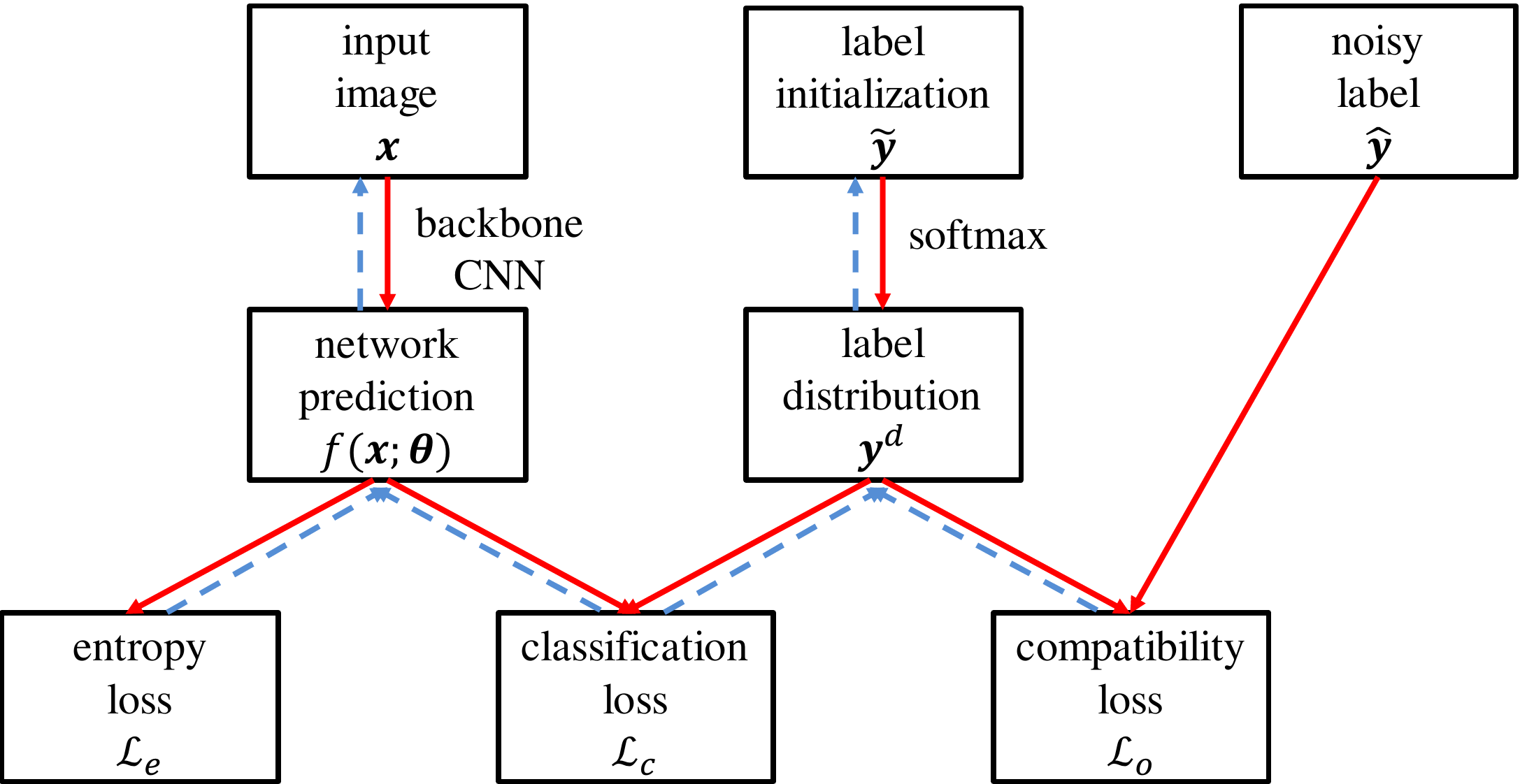}
\end{center}
   \caption{The PENCIL learning framework. We use label distributions $\vec{y}^d$ (which is the softmax transformed version of label initialization variables $\tilde{\vec{y}}$) to replace noisy labels $\hat{\vec{y}}$. The label distributions are updated in every iteration using three loss functions, among which the classification loss and compatibility loss updates $\vec{y}^d$ by requiring the label distributions produce both smooth models and not too distant from the noisy labels.}
\label{fig:1}
\end{figure}

Our label distribution $\vec{y}^d$ models the unknown noise-free label for $\vec{x}_i$. Hence, we need to estimate these distributions in our learning process. Let $\vec{X}$ and $\vec{Y}^d$ be the union of $\vec{x}_i$ and $\vec{y}_i^d$ (for all $1 \le i \le n$), respectively. Inspired by \cite{Tanaka2018CVPR_Daiki}, we let $\vec{Y}^d$ be part of the parameters that are to be updated in the back-propagation process. That is, PENCIL not only updates the network parameters $\vec{\theta}$ as in traditional networks, but also updates $\vec{Y}^d$ (i.e., $\vec{y}_i^d$) in every iteration. Therefore, we optimize both network parameters and label distributions as follows:
\begin{equation}
\min\limits_{\vec{\theta}, \vec{Y}^d} \mathcal{L}(\vec{\theta}, \vec{Y}^d|\vec{X})\label{6}
\end{equation}
The overall architecture of PENCIL is shown in Fig.~\ref{fig:1}.

In the PENCIL framework, three types of ``labels'' ($\vec{y}^d$, $\hat{\vec{y}}$ and $\tilde{\vec{y}}$) are involved. Label distribution $\vec{y}^d$ is updated by back-propagation. In the end, $\vec{y}^d$ will be a good estimate of the underlying unknown noise-free label (i.e., noise corrected label). $\tilde{\vec{y}}$ is a variable that assists $\vec{y}^d$ to be normalized to a probability distribution, by
\begin{align}
\vec{y}^d = \mathrm{softmax}(\tilde{\vec{y}}) \,.
\label{7}
\end{align}
Hence, $\tilde{\vec{y}}$ is not constrained and can be updated freely using back-propagation, but $\vec{y}^d$ is always a valid distribution.

The original noisy label $\hat{\vec{y}}$ does not directly impact the parameter ($\vec{\theta}$) learning. However, it is useful because we use it to indirectly initialize our label distribution $\vec{y}^d$. At the start of PENCIL, $\tilde{\vec{y}}$ is initialized by $\hat{\vec{y}}$ as follows:
\begin{equation}
\tilde{\vec{y}} = K \hat{\vec{y}} \,,
\label{8}
\end{equation}
where $K$ is a large constant ($K=10$ in our experiments), and hence from Eq.~\ref{7} we have $\vec{y}^d \approx \hat{\vec{y}} $ after this initialization.

\subsection{Compatibility loss}

The noisy label $\hat{\vec{y}}$ is also useful in PENCIL's loss computation. In fact, there are lots of (e.g., 80\% of) correct labels even in datasets with noisy labels. Therefore, we should not let the estimated label distribution $\vec{y}^d$ be completely different from those noisy labels $\hat{\vec{y}}$. 

We define a compatibility loss $\mathcal{L}_{o}(\hat{\vec{Y}},\vec{Y}^d)$ to enforce this requirement, as
\begin{align}
\mathcal{L}_{o}(\hat{\vec{Y}},\vec{Y}^d) = -\frac{1}{n}\sum_{i=1}^n\sum_{j=1}^c \hat{y}_{ij}\log y^d_{ij} \,,
\end{align}
which is a classic cross entropy loss between label distribution and noisy label. 

\subsection{Classification loss}\label{sec:class_loss}

The deviation between our label distribution $\vec{y}^d$ and the network prediction $f(\vec{x};\vec{\theta})$ guides how the network parameters $\vec{\theta}$ should be updated. In DLDL~\cite{Gao2017TIP_DLDL} and a  similar work~\cite{Tanaka2018CVPR_Daiki}, the classic KL-loss (Eq.~\ref{2}) is used to calculate the distance between these two distributions. However, we find that Eq.~\ref{2} works poorly in PENCIL and propose to use Eq.~\ref{4} instead, as a new classification loss (which we denote as $\mathcal{L}_c$).

Because we need to update the label distribution, we need to calculate $\frac{\partial \mathcal{L}_c}{\partial\vec{y}^d}$. If Eq.~\ref{2} is used as the classification loss $\mathcal{L}_c$, then
\begin{equation}
\frac{\partial \mathcal{L}_c}{\partial y_{ij}^d} = 1 + \sum_{j=1}^c \log\frac{y_{ij}^d}{f_j(\vec{x}_i; \vec{\theta})} \,.
\label{11}
\end{equation}

And, if we use Eq.~\ref{4} as $\mathcal{L}_c$, we have
\begin{equation}
\frac{\partial \mathcal{L}_c}{\partial y_{ij}^d} = -\sum_{j=1}^c \frac{f_j(\vec{x}_i; \vec{\theta})}{y_{ij}^d} \,.
\label{12}
\end{equation}

Then, we have the following observations for a fixed training example $i$ and any class index $j$.
\begin{itemize}
\item[Case 1] If the prediction $f_j(\vec{x}_i; \vec{\theta})$ is much larger than label distribution $y^d_{ij}$, Eq.~\ref{11} leads to a medium negative gradient (because of the $\log$), but Eq.~\ref{12} leads to a large negative gradient for updating $y_{ij}^d$. 

\item[Case 2] If $f_j(\vec{x}_i; \vec{\theta})$ is much smaller than $\vec{y}^d_j$, Eq.~\ref{11} leads to a medium positive gradient while Eq.~\ref{12} leads to a gradient which is almost zero.
\end{itemize}

Suppose for $\vec{x}_i$ the noisy label $\hat{y}_i$ is peaked at $j=3$ (i.e., $\hat{y}_{i,3}=1$) but the true label is 7. Thus, initially $y^d_{i,3}$ will be the peak in our label distribution $\vec{y}^d_i$. The internal smoothness inside the network may make the prediction $f(\vec{x}_i;\vec{\theta})$ to (correctly) peak at $j=7$. Hence, we have $f_7(\vec{x}_i;\vec{\theta}) \gg \hat{y}_{i,7}$ and $f_3(\vec{x}_i;\vec{\theta}) \ll \hat{y}_{i,3}$. Eq.~\ref{4} (Eq.~\ref{12}) will then (correctly) increase $y_{i,7}^d$ by a large amount, while Eq.~\ref{2} (Eq.~\ref{11}) will not (Case 1). Now consider the updating of $y^d_{i,3}$. Eq.~\ref{2} (Eq.~\ref{11}) will only decrease $y^d_{i,3}$ by a medium amount, and Eq.~\ref{4} (Eq.~\ref{12}) will keep $y^d_{i,3}$ almost intact (Case 2).

Combining these observations altogether, we believe that although the classic KL-loss (Eq.~\ref{2}) is a good fit for other applications, our proposed Eq.~\ref{4} is more suitable for correcting the noise in labels. Hence, we use the variant of KL-loss in Eq.~\ref{4} as our classification loss $\mathcal{L}_c$.

\subsection{Entropy loss}
Obviously, when the prediction $f(\vec{x};\vec{\theta})$ is the same as the label distribution $\vec{y}^d$, the network will stop updating. However, $f(\vec{x};\vec{\theta})$ tend to approach $\vec{y}^d$ fairly quickly, because label distributions are used as the supervision signal for learning network parameters $\vec{\theta}$. Following~\cite{Tanaka2018CVPR_Daiki}, we add an additional loss (regularization) term to avoid this problem. The entropy loss can force the network to peak at only one category rather than being flat because the one-hot distribution has the smallest possible entropy value. This property is advantageous for classification problems. The entropy loss is defined as 
\begin{align}
\mathcal{L}_e(f(\vec{x};\vec{\theta})) = -\frac{1}{n}\sum_{i=1}^n\sum_{j=1}^c f_j(\vec{x}; \vec{\theta})\log f_j(\vec{x}; \vec{\theta}) \,.
\end{align} 
At the same time, it also helps avoid the training from being stalled in our PENCIL framework, because the label distribution is not going to be a one-hot distribution and then $f(\vec{x};\vec{\theta})$ will be different from $\vec{y}^d$.

\subsection{The overall PENCIL framework}

With all components ready, the PENCIL loss function is
\begin{gather}
\mathcal{L} = \frac{1}{c}\mathcal{L}_c(f(\vec{x};\vec{\theta}),\vec{Y}^d)+\alpha \mathcal{L}_{o}(\hat{\vec{Y}},\vec{Y}^d)+\frac{\beta}{c} \mathcal{L}_e(f(\vec{x};\vec{\theta})) \,, \nonumber
\label{9}
\end{gather}
in which $\alpha$ and $\beta$ are two hyperparameters. Using this loss function and the PENCIL framework's architecture in Fig.~\ref{fig:1}, we can use \emph{any} deep neural network as the backbone network in Fig.~\ref{fig:1}, and then equip it with the PENCIL network to handle learning problems with noisy labels. The relationship between variables and loss functions are clearly visualized in Fig.~\ref{fig:1} as arrows. Forward computations are visualized by red solid arrows, while back-propagation computations are visualized as blue dashed arrows. The algorithmic description of the PENCIL framework is shown in Algorithm~\ref{A1}.

\begin{algorithm}[t]
\caption{The proposed PENCIL framework} 
\textbf{Input:} the noisy training set \{$\vec{x}_i$, $\hat{\vec{y}}_i$\} $(1 \le i \le n)$, and the number of training epochs $T$
\begin{algorithmic}[1]
\item initialize $\tilde{\vec{y}}_i$ $(1 \le i \le n)$ by Eq.~\ref{8}
\item $t \leftarrow 1$

\While {$t \le T$} 
 \State update $\vec{\theta}$ and $\vec{y}_i^d$ ($1 \le i \le n$) by forward computation and backward propagation in the mini-batch fashion using all $n$ training examples (i.e., to finish one epoch)
 \State $t \leftarrow t+1$
\EndWhile
\end{algorithmic}
\textbf{Output:} the trained network model $\vec{\theta}$, and the noise corrected labels $\vec{y}^d_i$ $(1 \le i \le n)$.
\label{A1}
\end{algorithm}

We want to add two notes about PENCIL. First, the error back-propagation process in PENCIL is pretty straightforward. For example, it can be done automatically in deep learning packages that support automatic gradient computation. Second, after the network has been fully trained (cf. Section~\ref{sec:exp}), those PENCIL-related components in Fig.~\ref{fig:1} are \emph{not needed} at all---the backbone network alone can perform prediction for future test examples.

Similar to \cite{Tanaka2018CVPR_Daiki}, we implement our PENCIL training through 3 steps.

\textbf{Backbone learning:}  We firstly train the backbone network with a large fixed learning rate from scratch without noise handling. As aforementioned, it is observed that when the learning rate is high, a DNN often does not overfit the label noise. Therefore, in this step, we use a fixed high learning rate with only the cross-entropy loss function in Eq.~\ref{1}. The resulted DNN is the backbone network in Fig.~\ref{fig:1}.

\textbf{PENCIL learning:} Then, we use the PENCIL framework to update both network parameters and label distributions. The learning rate is still a fixed high value. Therefore, the network will not overfit label noise and the label distributions will correct noise in the original labels. At the end of this step, we obtain a label distribution vector for every image. Algorithmic details are shown in Algorithm~\ref{A1}. Note that in practice we find that updating $\tilde{\vec{y}}$ requires a learning rate that is much larger than that used for updating other parameters. Because the overall learning rate is fixed in this step, we simply use one single hyperparameters $\lambda$ to update $\tilde{\vec{y}}$ (i.e., do not use PENCIL's overall learning rate), as
\begin{equation}
 \tilde{\vec{y}} \leftarrow \tilde{\vec{y}} - \lambda \frac{\partial \mathcal{L}}{\partial \tilde{\vec{y}}} \,.
\end{equation}

\textbf{Final fine-tuning:} Lastly, we use the learned label distributions to fine-tune the network using only the classification loss $\mathcal{L}_c$ (i.e., $\alpha=\beta=0$). In this step, the label distributions will not be updated and the learning rate will be gradually reduced as in common neural network training.

\section{Experiments} \label{sec:exp}
We tested the proposed PENCIL framework on both synthetic and real-world datasets: CIFAR-100~\cite{Krizhevsky2009TR_CIFAR}, CIFAR-10~\cite{Krizhevsky2009TR_CIFAR},  CUB-200~\cite{Wah2011CIT_CUB_200_2011} and Clothing1M~\cite{Xiao2015CVPR_Clothing1M}. All experiments were implemented using the PyTorch framework.

\subsection{Datasets}

\textbf{CIFAR-100:} Following~\cite{Zhang2018NIPS_Lq}, we retained 10\% of the training data as the validation set, and \emph{both} train and validation sets were noise contaminated. However, note that we did \emph{not} use the validation set in our method, because PENCIL \emph{does not need a validation set}. 

There are two types of noises: symmetric and asymmetric. Following~\cite{Zhang2018NIPS_Lq}, in the symmetric noise setup, label noise is uniformly distributed among all categories, and the label noise percentage is $r \in [0,1]$. For every example, if the correct label is $i$, then the noise-contaminated label has $1-r$ probability to remain correct, but has $r$ probability to be drawn uniformly from the $c$ labels. The asymmetric noise label was generated by flipping each class to the next class circularly with noise rate $r \in [0,1]$. 

\textbf{CIFAR-10:} Following~\cite{Tanaka2018CVPR_Daiki}, we retained 10\% of the CIFAR-10 training data as the validation set and modify the original correct labels to obtain different noisy label datasets. The setting for symmetric noise is the same as that in CIFAR-100. As for asymmetric noise, following \cite{Patrini2017CVPR_Forward} the noisy labels were generated by mapping \texttt{truck} $\rightarrow$ \texttt{automobile}, \texttt{bird}$\rightarrow$ \texttt{airplane}, \texttt{deer} $\rightarrow$ \texttt{horse} and \texttt{cat} $\leftrightarrow$ \texttt{dog} with probability $r$. These noise generation methods are in coincidence with confusions that often happen in the real world.

\textbf{Clothing1M:} Clothing1M is a large-scale dataset with noisy labels. It consists of more than one million images from 14 classes with many wrong labels. Images were obtained from several online shopping websites and labels were generated by their surrounding texts. The estimated noise level is roughly 40\%~\cite{Xiao2015CVPR_Clothing1M}. This dataset is seriously imbalanced and the label mistakes mostly happen between similar classes (i.e., asymmetric). There exist additional training, validation and test sets with 50k, 14k and 10k examples whose labels are believed to be clean, respectively.

\textbf{CUB-200:} We tested the robustness of our framework in a fine-grained classification dataset CUB-200. CUB-200 contains 11788 images of 200 species of birds, which is not considered to have the noisy label difficulty. Therefore, we tested our framework on this dataset to show that PENCIL is robust.  In addition, there is probably a small percentage of noisy labels in CUB-200~\cite{Wilber2015ICCV_NoisyLabelCUB}. It is interesting to observe whether PENCIL is robust and effective in such a dataset.

\subsection{Implementation details}\label{sec:setup}
 
Next, we describe more implementation details for each dataset.

\textbf{CIFAR-100:} We used ResNet-34~\cite{He2016CVPR_ResNet} as the backbone network for fair comparison with existing methods. The learning rate was $0.35$, $\alpha=0.1$, $\beta=0.4$, and $\lambda=10000$. Mean subtraction, horizontal random flip and $32 \times 32$ random crops after padding 4 pixels on each side were performed as data preprocessing and augmentation. We used SGD with $0.9$ momentum, a weight decay of $10^{-4}$, and batch size of $128$. Following~\cite{Tanaka2018CVPR_Daiki}, the epoch numbers for three steps were 70, 130 and 120, respectively. In the last step, we used the learning rate of $0.2$ and divided it by 10 after 40 and 80 epochs~\cite{Tanaka2018CVPR_Daiki}. All experiments on CIFAR-100 used the same settings as described above. In fact, we can obtain better results by further tuning the hyperparameters (e.g., as what we will soon introduce for CIFAR-10). However, we choose to use the same set of hyperparameters to demonstrate the robustness of our framework.

\textbf{CIFAR-10:} We used PreAct ResNet-32~\cite{He2016ECCV_ResnetV2} as the backbone network for fair comparison with existing methods. We used the same settings as those for CIFAR-100, except the overall learning rate, $\alpha$, $\beta$ and $\lambda$ hyperparameters. On CIFAR-10, these hyperparameters are shown in Table~\ref{tb1}. 

\begin{table}
\centering
\footnotesize
\caption{Hyperparameters for CIFAR-10 experiments. 3000 $\rightarrow$ 0 means that $\lambda$ decreases from 3000 to 0 linearly.}
\begin{tabular}{c|cccc}
\hline
\multicolumn{5}{c}{Symmetric Noise} \\
\hline   
noise rate (\%)& 	learning rate& 	$\alpha$& 	$\beta$& 	$\lambda$\\
\hline
10	& 	0.02	& 	0.1	& 	0.8	& 	200	\\
30	& 	0.03	& 	0.1	& 	0.8	& 	300	\\
50	& 	0.04	& 	0.1	& 	0.8	& 	400	\\
70	& 	0.08	& 	0.1	& 	0.8	& 	800	\\
90	& 	0.12	& 	0.1	& 	0.4	& 	1200\\
\hline
\hline 
\multicolumn{5}{c}{Asymmetric Noise}\\
\hline   
noise rate (\%)& 	learning rate& 	$\alpha$& 	$\beta$& 	$\lambda$\\
\hline
10	& 	0.06	& 	0.1	& 	0.4	& 	600	\\
20	& 	0.06	& 	0.1	& 	0.4	& 	600	\\
30	& 	0.06	& 	0.1	& 	0.4	& 	600	\\
40	& 	0.03	& 	0\phantom{.0} & 	0.4	& 	3000 $\rightarrow$ 0	\\
50	& 	0.03	& 	0\phantom{.0} & 	0.4	& 	4000 $\rightarrow$ 0\\
\hline
\end{tabular}
\label{tb1}
\end{table}

As shown in Table~\ref{tb1}, the learning rate increases as the noise rate increases for symmetric noise. This is reasonable, because when noise rate gets higher, we need stronger robustness and we can increase the learning rate to prevent our network from overfitting. And, when the noise rate is very high (e.g., 50\% asymmetric), there are too many noisy labels. Hence, we can remove the effect of noisy labels by removing $\mathcal{L}_o$ (i.e., set $\alpha$ to 0). At the same time, we require a large $\lambda$ to correct these noisy labels quickly. However, after a few epochs, the noisy labels were quickly corrected to a stable state (cf. Fig.~\ref{fig:2} and Fig.~\ref{fig:3}). Hence, we need to decrease $\lambda$ linearly to prevent wrong updates in later epochs.

\textbf{CUB-200:} On this dataset, we used ResNet-50~\cite{He2016CVPR_ResNet} pre-trained on ImageNet. Data preprocessing and augmentation is also applied, including performing mean subtraction, horizontal random flip, resizing the image to $256 \times 256$ and $224 \times 224$ random crops. We used SGD with $0.9$ momentum, a weight decay of $10^{-4}$, and batch size of $16$. The number of epochs for the three steps are 35, 65 and 60, respectively. The learning rate of the first and second step is $2 \times 10^{-3}$. In the last step, the learning rate is $10^{-3}$ and divided by 10 after 20 epochs and 40 epochs. $\beta$ is 0.8 and we reported results for different values of $\alpha$ and $\lambda$ as ablation studies.

\textbf{Clothing1M:} We used ResNet-50 pre-trained on ImageNet as the backbone network for fair comparison with existing methods. Data preprocessing and augmentation are the same as those in CUB-200. We used SGD with $0.9$ momentum, a weight decay of $10^{-3}$, and batch size of $32$. The epoch numbers of three steps are 5, 10 and 10, respectively. The first step learning rate is $1.6 \times 10^{-3}$ and the second step learning rate is $8 \times 10^{-4}$. The last step learning rate is $5 \times 10^{-4}$ and divided by 10 after 5 epochs. $\alpha = 0.08$, $\beta = 0.8$. In first 5 epochs of second step $\lambda = 3000$, and in last 5 epochs of second step $\lambda = 500$.

This dataset exists serious data imbalance. Therefore, we randomly selected a small balanced subset (using the noisy labels) to relieve the difficulty caused by imbalance. The small subset includes about 260k images and all classes have the same number of images. All our experiments on Clothing1M were done with this subset in this study. However, note that this subset is not truly balanced, because the labels are noisy.

\subsection{Results on CIFAR-100}
\begin{table*}
\centering
\footnotesize
\caption{Results on CIFAR-100. We report the average accuracy and standard deviation of 5 trials. $\#1$ to $\#5$ are quoted from \cite{Zhang2018NIPS_Lq}. PENCIL ($\#6$) is the result of last epoch (without using the validation set). The row with a star \textbf{*} ($\#2$) did not participate in comparison for fairness.} 
\begin{tabular}{@{\,}c@{\,}|c|cccc|cccc}
\hline
$\#$ & method & \multicolumn{4}{c|}{Symmetric Noise} & \multicolumn{4}{c}{Asymmetric Noise} \\
\hline
 & noise rate (\%) & 20 & 40 & 60 & 80 & 10 & 20 & 30 & 40 \\
\hline
1 & Cross Entropy Loss & 58.72$\pm$0.26& 48.20$\pm$0.65& 37.41$\pm$0.94& 18.10$\pm$0.82& 66.54$\pm$0.42& 59.20$\pm$0.18& 51.40$\pm$0.16& 42.74$\pm$0.61\\

2 & Forward $T$ \textbf{*}~\cite{Patrini2017CVPR_Forward}& {63.16$\pm$0.37}& {54.65$\pm$0.88}& {44.62$\pm$0.82}& {24.83$\pm$0.71}& {71.05$\pm$0.30}& {71.08$\pm$0.22}& {70.76$\pm$0.26}& {70.82$\pm$0.45}\\

3 & Forward $\hat{T}$~\cite{Patrini2017CVPR_Forward} & 39.19$\pm$2.61& 31.05$\pm$1.44& 19.12$\pm$1.95& \phantom{0}8.99$\pm$0.58& 45.96$\pm$1.21& 42.46$\pm$2.16& 38.13$\pm$2.97& 34.44$\pm$1.93\\ 

4 & $\mathcal{L}_q$~\cite{Zhang2018NIPS_Lq} & 66.81$\pm$0.42& 61.77$\pm$0.24& 53.16$\pm$0.78& 29.16$\pm$0.74& 68.36$\pm$0.42& 66.59$\pm$0.22& 61.45$\pm$0.26& 47.22$\pm$1.15\\

5 & Trunc $\mathcal{L}_q$~\cite{Zhang2018NIPS_Lq} & 67.61$\pm$0.18& 62.64$\pm$0.33& 54.04$\pm$0.56&\textbf{29.60$\pm$0.51}& 68.86$\pm$0.14& 66.59$\pm$0.23& 61.87$\pm$0.39& 47.66$\pm$0.69\\
\hline
6 & PENCIL ($last$) & \textbf{73.86$\pm$0.34}& \textbf{69.12$\pm$0.62}& \textbf{57.79$\pm$3.86} & fail & \textbf{75.93$\pm$0.20}& \textbf{74.70$\pm$0.56}& \textbf{72.52$\pm$0.38}&  \textbf{63.61$\pm$0.23}\\
\hline
\end{tabular}
\label{tb2}
\end{table*}

Firstly we tested PENCIL on  CIFAR-100. The results are shown in Table~\ref{tb2}. All dataset settings followed~\cite{Zhang2018NIPS_Lq}. The method ``Forward $T$~\cite{Patrini2017CVPR_Forward}'' used the groundtruth noise transition matrix (which is not available in real-world datasets), hence its numbers were not compared with other methods. Except for the 80\% symmetric noise case, PENCIL significantly outperformed previous methods in all symmetric and asymmetric noise cases. Even if ``Forward $T$'' used strong prior information which should not have been used, our PENCIL method still outperformed it in most cases. 

As for the 80\% symmetric noise case, it revealed a \emph{failure mode} of the proposed PENCIL method. When the noise rate is too high (e.g., 80\%), the correct labels only form a minority group and they are too weak to bootstrap the noise correction process. Hence, PENCIL tends to fail in such high noise rate problems. Fortunately, we hardly deal with such high noise rate in real-world applications. For example, the large scale real-world image dataset JFT300M~\cite{Sun2017ICCV_JFT300M} only includes about 20\% noisy labels.

We have intentionally chosen the same set of hyperparameters in all experiments on this dataset, and the results demonstrate the \emph{robustness} of our PENCIL framework to these hyperparameters. We can obtain better accuracy by using different hyperparameters for different noise rate and noise type, as shown in Table~\ref{tb1} on the CIFAR-10 dataset.

\subsection{Experiments on CIFAR-10}

Next, we evaluated the performance of our PENCIL framework on CIFAR-10. All the settings have been described in Section~\ref{sec:setup}. On the original noise-free CIFAR-10 dataset, the result of our backbone network (PreAct ResNet-32) is $94.05\%$. Our setup followed that in~\cite{Tanaka2018CVPR_Daiki}. However, results in~\cite{Tanaka2018CVPR_Daiki} used a prior knowledge (i.e., all categories have the same number of noise-free training examples), which should not be used. For fair comparison, we implemented the ``Tanaka \emph{et al.}~\cite{Tanaka2018CVPR_Daiki}'' method and in our implementation we did not use this prior knowledge.

Table~\ref{tb3} lists results of symmetric noise for CIFAR-10. In Table~\ref{tb3}, ``$best$'' denotes the test accuracy of the epoch where the validation accuracy was optimal and ``$last$'' denotes the test accuracy of the last epoch. As aforementioned, when the learning rate is small, the deep neural network's accuracy will decline because the network memorizes all the (noisy) labels, i.e., the network is overfitting. As shown in row $\#1$, the traditional neural network using the classic cross entropy loss is heavily affected by this difficulty. Its $best$-epoch test accuracy was significantly better than that of the $last$-epoch one. And, as the noise rate increased, the gap was even larger because the overfitting to noise became more serious as expected. On the contrary, our method and the Tanaka~\textit{et~al.}~\cite{Tanaka2018CVPR_Daiki} did not have obvious accuracy drop between $best$- and $last$-epochs. Therefore, the proposed PENCIL method has strong robustness. As for the test set accuracy, PENCIL had a clear advantage than competing methods in Table~\ref{tb3}. The winning gap became especially apparent when the noise rate increased to larger values. For example, when the noise rate was 90\%, PENCIL obtained roughly 7\% higher accuracy than that of Tanaka~\textit{et~al.} and 10\% higher than that of cross entropy.

\begin{table}
\centering
\footnotesize
\caption{Test accuracy on CIFAR-10 with symmetric noise. We reported the average result of 5 trials. All results in this table were based on our own implementation.}
\setlength{\tabcolsep}{3.5pt}
\begin{tabular}{c|c|c|ccccc}
\hline
$\#$ &method	&		\multicolumn{6}{c}{Symmetric Noise}\\	
\hline
&noise rate (\%)	& &	10	&	30	&	50	&	70	&	90	\\
\hline
\multirow{2}{*}{1}&\multirow{2}{*}{Cross Entropy Loss}	&	$best$	&	91.66	&	89.00	&	85.15	&	78.09	&	50.74	\\
&&$last$	&	88.43	&	72.78	&	53.11	&	33.32	&	16.30	\\
\hline
\multirow{2}{*}{2}&\multirow{2}{*}{Tanaka~\textit{et~al.}~\cite{Tanaka2018CVPR_Daiki}}&	$best$	&	93.23	&	91.23	&	88.50	&	84.51	&	54.36	\\
&&	$last$	&	93.23	&	91.22	&	88.51	&	84.59	&	53.49	\\
\hline
\multirow{2}{*}{3}&\multirow{2}{*}{PENCIL}&	$best$	&	{93.26}	&	{92.09}	&	{90.29}	&	87.10	&\textbf{61.21}	\\
&&	$last$	&\textbf{93.28}	&\textbf{92.24}	&\textbf{90.36}	&\textbf{87.18}	&	60.80	\\
\hline
\end{tabular}
\label{tb3}
\end{table}

Table~\ref{tb4} lists results of asymmetric noise for CIFAR-10. In terms of robustness, methods shown in row $\#1$, $\#2$ and $\#3$ had the overfitting problem and their test accuracies had large gaps between the $best$- and $last$-epochs. The Tanaka~\textit{et~al.} method experienced the same issue when the noise rate was high (50\%), but was robust in other cases. Our PENCIL method, however, remained robust throughout all the experiments. 

\begin{table}
\centering
\footnotesize
\caption{Test accuracy on CIFAR-10 with asymmetric noise. We reported the average result of 5 trials. Rows $\#1$, $\#4$ and $\#5$ were based on our own implementation. Rows $\#2$ and $\#3$ were quoted from \cite{Tanaka2018CVPR_Daiki}. The methods marked with a ``*'' used additional information that should not be used, and need to be excluded in a fair comparison.}
\setlength{\tabcolsep}{3.5pt}
\begin{tabular}{c|c|c|ccccc}
\hline
$\#$ & method	&		\multicolumn{6}{c}{Asymmetric Noise}\\	
\hline
&noise rate (\%)	& &	10	&	20	&	30	&	40	&	50	\\
\hline
\multirow{2}{*}{1}&\multirow{2}{*}{Cross Entropy Loss}	&$best$	&	91.09	&	89.94	&	88.78	&	87.78	&	77.79	\\
&&$last$	&	85.24	&	80.74	&	76.09	&	76.12	&	71.05	\\	
\hline
\multirow{2}{*}{2}&\multirow{2}{*}{Forward $T$ *~\cite{Patrini2017CVPR_Forward}}	&$best$	&	92.4\phantom{0}	&	91.4\phantom{0}	&	91.0\phantom{0}	&	90.3\phantom{0}	&	83.8\phantom{0}	\\
&&$last$	&	91.7\phantom{0}	&	89.7\phantom{0}	&	88.0\phantom{0}	&	86.4\phantom{0}	&	80.9\phantom{0}	\\
\hline
\multirow{2}{*}{3}&\multirow{2}{*}{CNN-CRF *~\cite{Vahdat2017NIPS_CNNCRF}}	&	$best$	&	92.0\phantom{0}	&	91.5\phantom{0}	&	90.7\phantom{0}	&	89.5\phantom{0}	&	84.0\phantom{0}	\\
&&	$last$	&	90.3\phantom{0}	&	86.6\phantom{0}	&	83.6\phantom{0}	&	79.7\phantom{0}	&	76.4\phantom{0}	\\
\hline
\multirow{2}{*}{4}&\multirow{2}{*}{Tanaka~\textit{et~al.}~\cite{Tanaka2018CVPR_Daiki}}&	$best$	&	92.53	&	91.89	&	91.10	&	91.48	&	75.81	\\
&&	$last$	&	92.64	&	91.92	&	91.18	&	\textbf{91.55}	&	68.35	\\
\hline
\multirow{2}{*}{5}&\multirow{2}{*}{PENCIL}&	$best$	&	93.00	&	\textbf{92.43}	&	\textbf{91.84}	&	91.01	&	\textbf{80.51}	\\
&&	$last$	&	\textbf{93.04}	&\textbf{92.43}	&	91.80	&	91.16	&	80.06	\\
\hline
\end{tabular}
\label{tb4}
\end{table}

The Forward~\cite{Patrini2017CVPR_Forward} and CNN-CRF~\cite{Vahdat2017NIPS_CNNCRF} methods both require the ground-truth noise transition matrix, which is hardly available in applications. Our method does not require any prior information about noise labels. Table~\ref{tb4} shows that PENCIL has been robust and is the overall accuracy winner on CIFAR-10.

\begin{figure}
 \centering
 \includegraphics[width=0.7\linewidth]{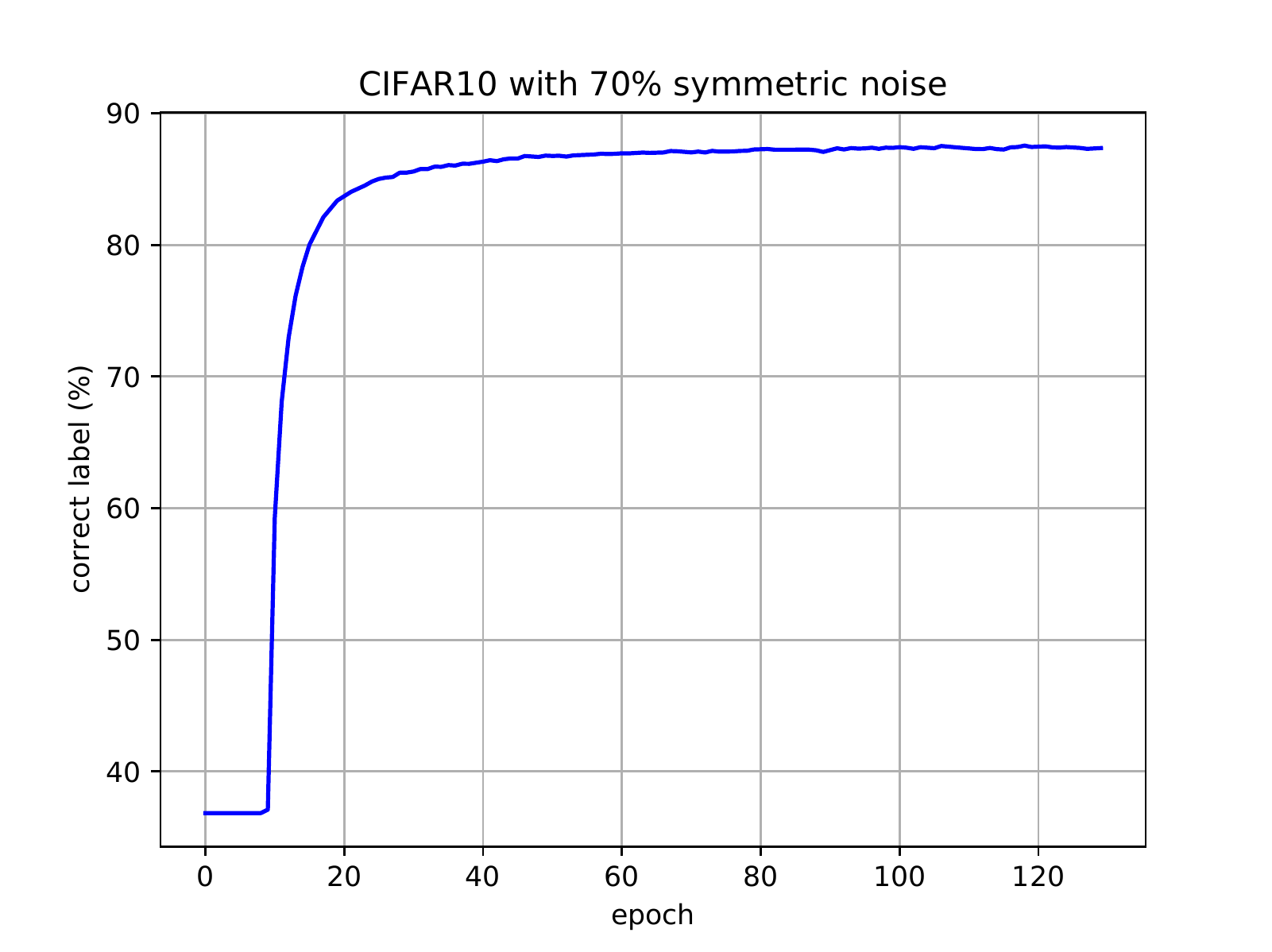}
 \caption{Correct labels on CIFAR-10 with 70\% symmetric noise.}
 \label{fig:2}
\end{figure}

\begin{figure}
 \centering
 \includegraphics[width=0.7\linewidth]{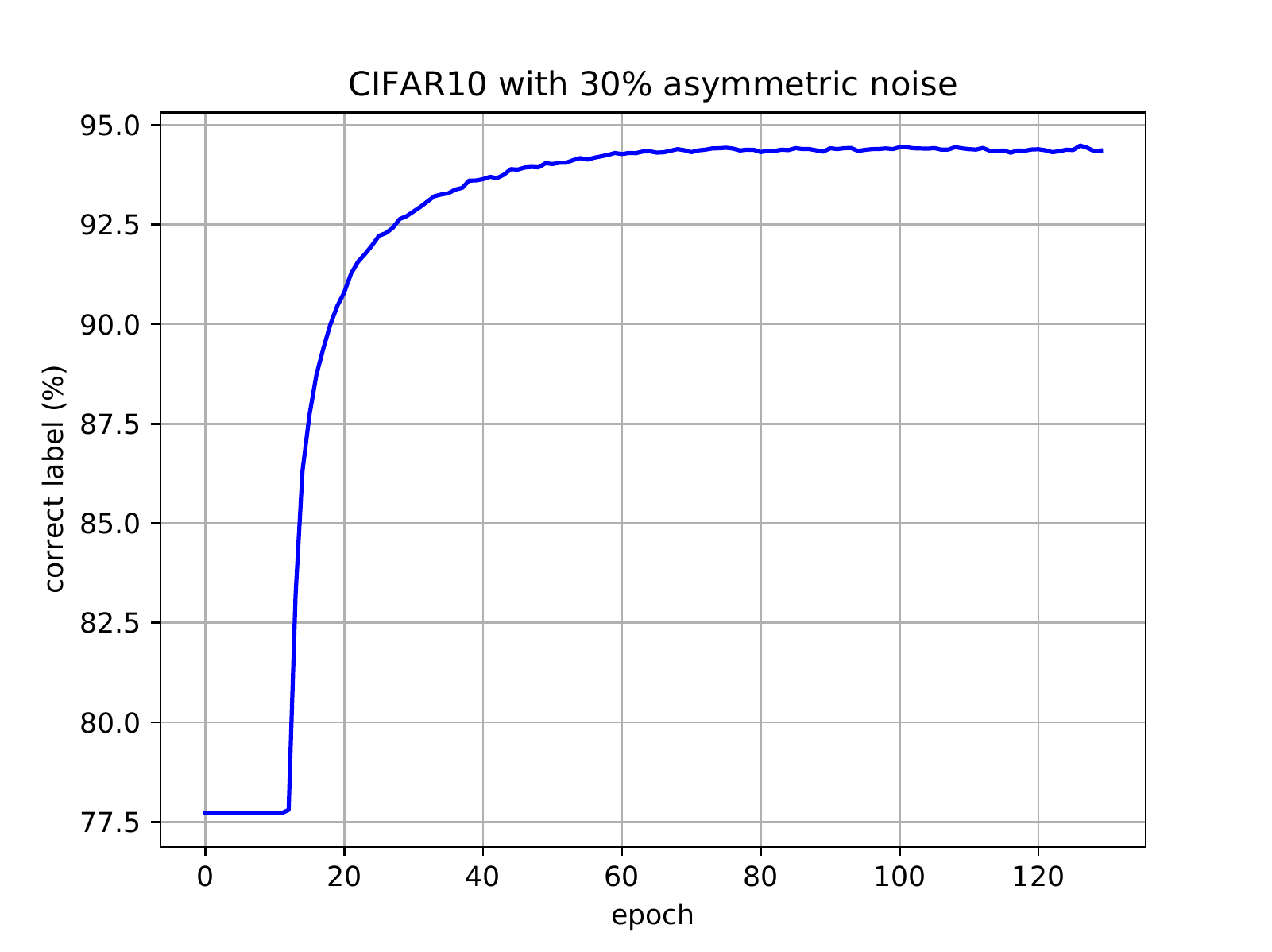}
 \caption{Correct labels on CIFAR-10 with 30\% asymmetric noise.}
 \label{fig:3}
\end{figure}

We recorded the number of correct labels in PENCIL's second step. In a label distribution vector, the category corresponding to the maximum value in the probability distribution was identified as the label estimated by PENCIL. If this label was the same as the noise-free groundtruth label, we say it was correct. The results for 70\% symmetric and 30\% asymmetric noise on CIFAR-10 are shown in Fig.~\ref{fig:2} and Fig.~\ref{fig:3}, respectively. We can observe that PENCIL effectively and stably estimated correct labels for most examples even with high noise rates. For example, with 70\% symmetric noise rate, originally only about 16000 labels were correct, but after PENCIL's learning process there are about 39000 correct labels.

\subsection{Experiments on CUB-200}

We performed additional experiments on CUB-200 with different hyperparameters $\alpha$ and $\lambda$. This dataset is generally considered to contain no or only few noisy labels. Therefore, we use it to further test the robustness of PENCIL on problems not affected by noisy labels.

The results are listed in Table~\ref{tb5}. Row $\#1$ is the baseline (classic method) and rows $\#2$ to $\#7$ are PENCIL results. For a wide range of $\alpha$ and $\lambda$ values, PENCIL consistently exhibited competitive results (i.e., without obvious degradation). Furthermore, we observed the final label distributions, and the maximum values of all label distributions are correct (i.e., same as the correct labels). This observation shows that PENCIL works robustly in clean datasets, too.

In the settings of rows $\#4$ to $\#7$, PENCIL achieved higher accuracy than the baseline. In particular, row $\#4$ is 0.71\% higher. A small percentage of label noise may exist in this dataset~\cite{Wilber2015ICCV_NoisyLabelCUB}. Our hypothesis is that by replacing the original one-hot label with probabilistic modeling in PENCIL, we obtained better robustness and consequently a small edge in accuracy.

\begin{table}
\footnotesize
\centering
\caption{Test accuracy on CUB-200 with different hyperparameters. The accuracy of PENCIL does not decline in standard datasets with clean labels.}
\begin{tabular}{c|p{1.5cm}<{\centering}p{1.5cm}<{\centering}|c}
\hline
$\#$ & \multicolumn{2}{c|}{method} & Test Accuracy (\%)\\
\hline
1& \multicolumn{2}{c|}{Cross Entropy Loss} & 81.93\\
\hline
\hline
&\multicolumn{2}{c|}{PENCIL}&\\
\hline
& $\lambda$& $\alpha$ & \\
\hline
2&1000&0\phantom{.0}&81.91\\
\hline
3&2000&0\phantom{.0}&81.84\\
\hline
4&3000&0\phantom{.0}&\textbf{82.64}\\
\hline
5&1000&0.1&82.09\\
\hline
6&2000&0.1&82.21\\
\hline
7&3000&0.1&82.22\\
\hline
\end{tabular}
\label{tb5}
\end{table}

\subsection{Experiments on Clothing1M}
Finally, we tested PENCIL on Clothing1M, which is a real-world noisy label dataset. It includes a lot of unknown structure (asymmetric) noise.

The results are shown in Table~\ref{tb6}. All results are $best$ test accuracy. Rows $\#1$ and $\#2$ were quoted from \cite{Patrini2017CVPR_Forward}, and row $\#3$ was reported in \cite{Tanaka2018CVPR_Daiki}. Although these baseline models were trained on the whole Clothing1M training set, our PENCIL used a randomly sampled pseudo-balanced subset, including about 260k images. The backbone network was ResNet-50 for all methods.

\begin{table}
\caption{Test accuracy on the Clothing1M dataset. Rows $\#1$ and $\#2$ were quoted from \cite{{Patrini2017CVPR_Forward}} and $\#3$ was quoted from \cite{Tanaka2018CVPR_Daiki}. These baseline methods used the complete Clothing1M training data, but our method only used a small pseudo-balanced subset (i.e., balanced in terms of noisy labels). Our method achieved state-of-the-art result in this real-world dataset.}
\footnotesize
\centering
\begin{tabular}{c|c|c}
\hline
$\#$&method & Test Accuracy (\%)\\
\hline
1& Cross Entropy Loss & 68.94\\
\hline
2& Forward~\cite{Patrini2017CVPR_Forward} & 69.84\\
\hline
3& Tanaka~\textit{et~al.}~\cite{Tanaka2018CVPR_Daiki} & 72.16\\
\hline
4& PENCIL & \textbf{73.49}\\
\hline
\end{tabular}
\label{tb6}
\end{table}

In Table~\ref{tb6}, only noisy labeled examples were used (i.e., without using the clean training subset). The Forward~\cite{Patrini2017CVPR_Forward} method required the ground-truth noise transition matrix, which is not available. Hence, it used an estimated matrix instead. The Tanaka~\textit{et~al.}~\cite{Tanaka2018CVPR_Daiki} method used the distribution of noisy labels to relieve the imbalanced problem. In our PENCIL method, we did not use any extra prior information. PENCIL achieved 1.33\% higher accuracy than that of Tanaka~\textit{et~al.}~\cite{Tanaka2018CVPR_Daiki}, 3.65\% higher than Forward~\cite{Patrini2017CVPR_Forward} and 4.55\% than cross entropy.

\section{Conclusion}
We proposed a framework named PENCIL to solve the noisy label problem. PENCIL adopted label probability distributions to supervise network learning and to update these distributions through back-propagation end-to-end in every epoch. We proposed a KL-loss, which is different from previous methods but is robust for noisy label handling. The proposed PENCIL framework is end-to-end and independent of the backbone network structure, thus it is easy to deploy.

We tested PENCIL with synthetic label noise on CIFAR-100 and CIFAR-10 with different noise types and noise rates, and outperformed current state-of-the-art methods by large margins. We also experimented on CUB-200, which is considered to be noise free. The results show that PENCIL is robust for different datasets and hyperparameters. Lastly, we tested PENCIL on the real-world large scale label noise dataset Clothing1M. On this dataset, we achieved 1.33\% higher accuracy than previous state-of-the-art.


{\small
\bibliographystyle{ieee_fullname}
\bibliography{PENCIL}
}

\end{document}